\documentclass{article}
\usepackage{ijcai24}

\usepackage{times}
\usepackage{soul}
\usepackage{url}
\usepackage[hidelinks]{hyperref}
\usepackage[utf8]{inputenc}
\usepackage[small]{caption}
\usepackage{graphicx}
\usepackage{amsmath}
\usepackage{amsthm}
\usepackage{amssymb}
\usepackage{booktabs}
\usepackage{algorithm}
\usepackage{algorithmic}
\usepackage[switch]{lineno}
\usepackage[table,xcdraw,dvipsnames]{xcolor}
\usepackage{multirow}
\usepackage{algorithm}
\usepackage{algorithmic}
\usepackage{pifont}
\usepackage{array}
\usepackage{dsfont}
\newcommand{\cmark}{\ding{51}}%
\newcommand{\xmark}{\ding{55}}

\newcolumntype{L}[1]{>{\raggedright\arraybackslash}p{#1}}
\newcolumntype{C}[1]{>{\centering\arraybackslash}p{#1}}
\newcolumntype{R}[1]{>{\raggedleft\arraybackslash}p{#1}}
\title{Learning Robust Classifiers with Self-Guided Spurious Correlation Mitigation}

\author{
Guangtao Zheng
\and
Wenqian Ye\And
Aidong Zhang
\affiliations
University of Virginia
\emails
\{gz5hp, pvc7hs, aidong\}@virginia.edu
}
\begin{document}

\maketitle

\begin{abstract}
Deep neural classifiers tend to rely on spurious correlations between spurious attributes of inputs and targets to make predictions, which could jeopardize their generalization capability. Training classifiers robust to spurious correlations typically relies on annotations of spurious correlations in data, which are often expensive to get. In this paper, we tackle an annotation-free setting and propose a self-guided spurious correlation mitigation framework. 
Our framework automatically constructs fine-grained training labels tailored for a classifier obtained with empirical risk minimization to improve its robustness against spurious correlations. The fine-grained training labels are formulated with different prediction behaviors of the classifier identified in a novel spuriousness embedding space. We construct the space with automatically detected conceptual attributes and a novel spuriousness metric which measures how likely a class-attribute correlation is exploited for predictions. 
We demonstrate that training the classifier to distinguish different prediction behaviors reduces its reliance on spurious correlations without knowing them a priori and outperforms prior methods on five real-world datasets.

\end{abstract}

\section{Introduction}
Deep neural classifiers have shown strong empirical performance in many application areas. However, some of the high performance may be credited to their strong reliance on spurious correlations \cite{sagawa2019distributionally,nam2020learning,nam2022spread}, which are brittle associations between non-essential spurious attributes of inputs and the corresponding targets in many real-world datasets. For example, a deep neural classifier trained with empirical risk minimization (ERM) can achieve a high accuracy of predicting cow by just detecting the \texttt{grassland} attribute of images, given that cow correlates with \texttt{grassland} in most images. However, the correlation is spurious as the \texttt{grassland} attribute is not essential for the class cow, and the classifier exhibits severe performance degradation on images showing a cow at a beach \cite{beery2018recognition,geirhos2020shortcut}.

Mitigating the reliance on spurious correlations is crucial for obtaining robust models. Existing methods typically assume that spurious correlations are known (1) fully in both the training and validation data for model training and selection \cite{sagawa2020investigation,deng2023robust} or (2) only in the validation data for model selection \cite{liu2021just,creager2021environment,kirichenko2022last,izmailov2022feature}. However, obtaining annotations of spurious correlations  typically requires expert knowledge and human supervision, which is a significant barrier in practice.

In this paper,  we tackle the setting where annotations of spurious correlations are not available. To train a classifier robust to spurious correlations without knowing them, we propose a novel self-guided spurious correlation mitigation framework that automatically detects and analyzes the classifier’s reliance on spurious correlations and relabels training data tailored for spurious correlation mitigation.

Our framework exploits the classifier's reliance on individual attributes contained in multiple training samples. To this end, we  first propose to automatically detect all possible attributes from a target dataset based on a pre-trained vision-language model (VLM). The VLM learns the mapping between real-world images and their text descriptions. Thus, it is convenient to extract informative words from the descriptions as the attributes which summarize similar input features that could be exploited by a classifier for predictions.

The detected attributes and class labels formulate all possible correlations that the classifier might exploit for predictions. The classifier may exploit some of the correlations for predictions and be invariant to others. To measure the classifier’s reliance on these class-attribute correlations or their degrees of spuriousness, we propose a spuriousness metric to quantify how likely the classifier relies on the correlations in a set of data. Given a class-attribute correlation, a large value of the metric shows the classifier’s strong reliance on the correlation and a significant impact of the correlation on the classifier's performance. 

With the spuriousness values for all the possible correlations, the prediction behaviors of the classifier on samples naturally emerge. The spuriousness values of all the class-attribute correlations relevant to a sample collectively characterize the classifier’s prediction behavior on the sample, i.e., how likely those correlations are exploited by the classifier for predicting the class label of the sample. Therefore, we can discover diverse prediction patterns of the classifier. For example, some of the prediction behaviors are frequently used on many samples while some are not.

To mitigate the classifier's reliance on spurious correlations, we demonstrate to the classifier that multiple prediction behaviors for samples in the same class are different but should lead to the same class label. In this way, the classifier is not only aware of multiple attributes leading to the same class, but also encouraged to discover more robust features for predictions. To achieve this, we relabel the training data with fine-grained labels so that the classifier is trained to distinguish samples from the same class with different prediction behaviors. 
Moreover, considering the imbalanced sample distributions over different prediction behaviors, we adopt a balanced sampling approach to train the adapted classifier for improved robustness against spurious correlations.

We consolidate the detection and mitigation methods in an iterative learning procedure since it is possible that mitigating certain spurious prediction behaviors may increase the reliance on others.   
Our method, termed as \textit{Learning beyond Classes (LBC)}, has the following \textbf{contributions}:
\begin{itemize}
    \item We completely remove the spurious correlation annotation barrier for learning a robust classifier by proposing an automatic detection method that exploits the prior knowledge in a pre-trained vision-language model.
    \item  We mitigate a classifier's reliance on spurious correlations by diversifying its outputs to recognize different prediction behaviors and balancing its training data.
    \item  Our method debiases a biased classifier with a new self-guided procedure of iteratively identifying and mitigating the classifier's spurious prediction behaviors. We demonstrate that LBC achieves the best performance on five real-world datasets where spurious correlations are unknown or unavailable.
\end{itemize}

\section{Related Work}
\textbf{Spurious correlation detection.} Spurious correlations are biases in data that could be harmful to a model's generalization. Discovering spurious correlations typically requires domain knowledge \cite{clark2019don,nauta2021uncovering} and human annotations \cite{nushi2018towards,zhang2018manifold}. Previous works discover that object backgrounds \cite{xiao2021noise} and image texture \cite{geirhos2018imagenettrained} could be spuriously correlated with target classes and severely bias the predictions of deep learning models. See \citeauthor{geirhos2020shortcut} for an overview. Recent works \cite{plumb2022finding,abid2022meaningfully} use model explanation methods to detect spurious features.  Neurons in the penultimate layer of a robust model are also exploited for spurious feature detection with limited human supervision \cite{singla2021salient,neuhaus2022spurious}. A pre-defined concept bank  \cite{wu2023discover} is also leveraged as an auxiliary knowledge base for spurious feature detection. In contrast, we propose a fully automatic spurious attribute detection method that exploits the prior knowledge in pre-trained vision-language models and extracts attributes in interpretable text format.

\noindent\textbf{Mitigating spurious correlations.} There is a growing number of works on mitigating the impact of spurious correlations. Typically, these works consider the problem in the context of group labels, which partition a dataset into groups of samples with the same class labels and spurious attributes.
\textit{When group labels are known}, balancing the sizes of groups \cite{cui2019class,he2009learning}, 
upweighting groups that do not have the specified spurious correlations \cite{byrd2019effect}, and optimizing the worst-group performance \cite{sagawa2019distributionally} are shown to be effective. \textit{When group labels are absent}, several works aim to infer group labels, including identifying misclassified samples \cite{liu2021just}, clustering hidden representations \cite{pmlr-v162-zhang22z}, invariant learning \cite{creager2021environment}, and training a group label estimator \cite{nam2022spread}.
\citeauthor{kirichenko2022last} uses a part of balanced validation data to retrain the last layer of a model. All the methods require group labels of the validation data for model selection, which is a strong assumption in practice. A recent work \cite{asgari2022masktune} uses masked data to mitigate the impact of spurious correlations without knowing them. Our method automatically detects spurious attributes in data and mitigates the reliance on them by constructing fine-grained classification tasks based our novel spuriousness metric. Another parallel line of works is to use data augmentation, such as mixup \cite{zhang2018mixup,han2022umix,wu2023discover} or selective augmentation \cite{yao2022improving}, to mitigate spurious bias in model training.

\begin{figure*}[t]
    \centering
    \includegraphics[width=\linewidth]{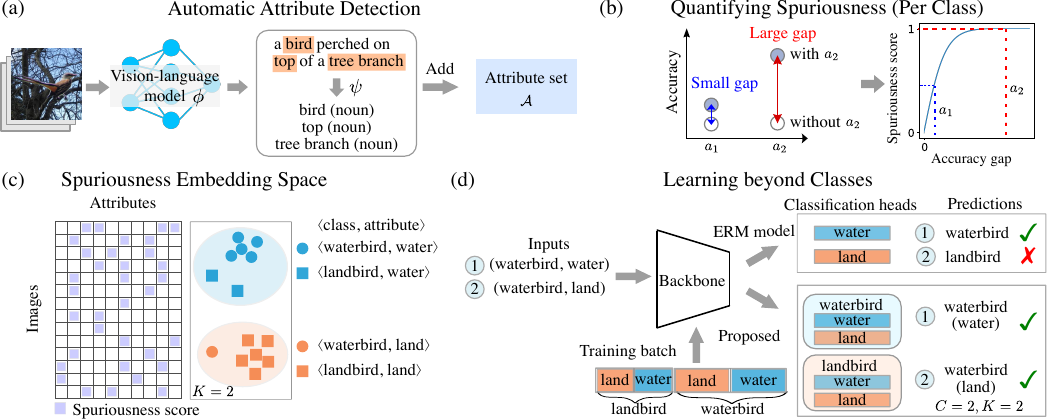}
    \caption{Method overview. (a) Detecting attributes with a pre-trained VLM. (b) Quantifying the spuriousness of correlations between classes and detected attributes. (c) Clustering in the spuriousness embedding space for relabeling  the training data. (d) Diversifying the outputs of the classifier and training the classifier with balanced training data.}
    \label{fig:method-overview}
\end{figure*}
\section{Problem Formulation}\label{sec:problem-formulation}
We focus on an ERM-trained classifier $f_{\theta}$ with parameters $\theta$ learned from a training dataset $\mathcal{D}_{tr}=\{(x_n,y_n)\}_{n=1}^N$ of $N$ pairs of image $x_n\in\mathcal{X}$ and label $y_n\in\mathcal{C}$, i.e.,
\begin{equation}\label{eq:erm-objective}
    \theta = \arg \min_{\theta'}\mathbb{E}_{(x,y)\in \mathcal{D}_{\text{tr}}}\ell(f_{\theta'}(x),y),
\end{equation}
where $\mathcal{X}$ is the set of all input samples, $\mathcal{C}$ is the set of classes, and $\ell(\cdot,\cdot)$ is the cross-entropy loss function. Before we introduce the problem regarding the classifier, we describe the following concepts we use in the paper.

\noindent\textbf{Group labels.} A group label $\langle c,a\rangle$ is a fine-grained label that uniquely annotates a group of samples which are in the class $c$ and have the attribute $a$.

\noindent\textbf{Spurious attributes.} A spurious attribute $a$ describes non-essential conceptual features of inputs. For example, $a=\texttt{water}$  may describe different water backgrounds in images.

\noindent\textbf{Spurious correlations.} A spurious correlation is the association between a spurious attribute $a$ and a class $c$ that exists \textit{only} in some samples of class $c$. We use $\langle c,a\rangle$ to denote both a group label and a spurious correlation if $a$ is spurious.

A real-world dataset can be partitioned into several groups with unique group labels. If the groups are imbalanced in size, the classifier $f_{\theta}$ typically will exploit the spurious correlations in the majority groups for predictions and ignore those in the minority groups. This exposes a robustness issue of the classifier, especially when it is deployed in a setting where data distributions are shifted towards the minority groups.

Existing problem formulations assume the availability of group labels in both the training and validation data for model training and selection, respectively, or in the validation data only.  In this paper, we consider the novel setting where \textbf{no group labels are available}, which completely removes the barrier of human annotation in existing methods.

\section{Methodology}

\subsection{Automatic Spurious Correlation Detection}\label{sec:spurious-detection}
We propose an automatic and scalable method to discover all potential spurious correlations in a target dataset. Our method exploits the prior knowledge in a pre-trained vision language model (VLM) and consists of automatic attribute detection and quantifying the spuriousness of correlations between detected attributes and classes.

\subsubsection{Automatic Attribute Detection}
The detection procedure has the following two steps.

\noindent\textit{1. Generating Text Descriptions.} We use a pre-trained VLM $\phi$ to generate text descriptions of images without human supervisions. Since the model is general-purposed and is not fine-tuned on target datasets, it can discover general objects and patterns. For example, in Figure \ref{fig:method-overview}(a), besides the class object \texttt{bird}, the model also detects \texttt{tree branch}.  
 
\noindent\textit{2. Extracting Informative Words}. We detect attributes by identifying nouns and adjectives from text descriptions as these types of words are informative in representing objects and patterns in images. We use an automatic procedure $\psi$ (Section \ref{sec:experiment}) to extract these informative words from the text descriptions obtained in the first step.  For example, we extract \texttt{bird},  \texttt{top}, and \texttt{tree branch} from the description shown in Figure \ref{fig:method-overview}(a). We add the detected attributes to a set $\mathcal{A}$ as the set of all attributes detected from the images.

\noindent\textbf{Remark:} The VLM $\phi$ is only used for the one-time data preparation, i.e., detecting attributes from the training data; it does not involve in predictions. The inductive bias in $\phi$ only affects what attributes are detected.
For example, it may inaccurately describe a lemon on a tree as a yellow bird (not self-explanatory). It is our algorithmic design introduced later that effectively mitigates a classifier's reliance on spurious correlations,  even though some extracted words may not be self-explanatory in representing certain kinds of features.

\subsubsection{Quantifying Spuriousness}
A detected attribute in $\mathcal{A}$ may form a vague spurious correlation with a class, since it may not exist in images or some attributes are not spurious and represent class objects. More importantly, not all correlations are exploited by a classifier for predictions. To quantify the likelihood of a class-attribute correlation being spurious and used by a classifier, i.e., \textit{spuriousness} of the correlation, we propose a novel metric, termed \textit{spuriousness score}, that unifies the above cases.

Our spuriousness score for $\langle c,a\rangle$ is motivated by the observation that the classifier $f_{\theta}$ will generalize poorly on samples of class $c$ without the attribute $a$ if $f_{\theta}$ excessively relies on $a$ for predicting the class $c$. Intuitively, the score will be higher if $f_{\theta}$ has a larger performance discrepancy on images with and without $a$ and vice versa. We formally define our spuriousness score for $\langle c,a\rangle$ as follows.

\noindent\textbf{Definition 1 (Spuriousness score):} Given a class $c\in\mathcal{C}$, an attribute $a\in\mathcal{A}$, and a classifier $f_{\theta}$ trained on $\mathcal{D}_{tr}$, the spuriousness of $\langle c,a \rangle$ is calculated as follows,
\begin{equation}\label{eq:spuriousness-score}
    \gamma(a,c;f_{\theta},\mathcal{D}_{tr})= \tanh\Big({\text{Abs}(\log\frac{M(\mathcal{D}_{tr}^{\langle c,a \rangle};f_{\theta})}{M(\mathcal{D}_{tr}^{\langle c,\hat{a} \rangle};f_{\theta})})}\Big),
\end{equation}
with $\gamma(a,c;f_{\theta},\mathcal{D}_{tr})=0$ when $\mathcal{D}_{tr}^{\langle c,\hat{a} \rangle}=\emptyset$ or $\mathcal{D}_{tr}^{\langle c,a \rangle}=\emptyset$, where $\mathcal{D}_{tr}^{\langle c,a \rangle}\subset\mathcal{D}_{tr}$ denotes the subset of all training samples from class $c$ \textbf{with} the attribute $a$, $\mathcal{D}_{tr}^{\langle c,\hat{a} \rangle}\subset\mathcal{D}_{tr}$ denotes the subset of all training samples from class $c$ \textbf{without} the attribute $a$, $M(\cdot;f_{\theta})$ denotes the classification accuracy of $f_{\theta}$ on a given set of samples, and Abs($\cdot$) denotes taking the absolute value. 
% Moreover, the division can produce larger values than the simple difference between the two accuracies, making different correlations more distinctive. 
Using $\log(\cdot)$ avoids encountering extreme values from the division, and $\tanh(\text{Abs}(\cdot))$ bounds the score in the range from 0 to 1.  Figure \ref{fig:method-overview}(b) gives an example. Other designs of the score are explored in Appendix.
 
\noindent \textbf{Remark:} The score depends on both the training data and the classifier. Consider the two corner cases: when $\mathcal{D}_{tr}^{\langle c,\hat{a} \rangle}=\emptyset$, $a$ is always associated with $c$ in $\mathcal{D}_{tr}$, e.g., $a$ is a class object; and when $\mathcal{D}_{tr}^{\langle c,a \rangle}=\emptyset$, $a$ and $c$ are not associated, i.e., no samples correspond to the association between $a$ and $c$. In both cases, the spuriousness score is zero as they do not fit in our description of spurious correlations. We exclude such attributes from $\mathcal{A}$. 
Moreover, the $\text{Abs}(\cdot)$ operator implies that when the fraction in Eq. \eqref{eq:spuriousness-score} is lower than 1, i.e., $a$ leads to incorrect predictions other than $c$ and results in a small nominator, $\langle c,a \rangle$ still has a high spuriousness score as it also represents a robustness pitfall.

With the spuriousness score, we can quantify the impact of the correlation between the detected attribute $a$ and the class $c$ on the performance of the classifier $f_{\theta}$. The lower the spuriousness score, the less $f_{\theta}$ relies on attribute $a$ in its prediction, thus the more robust the classifier is.

\subsection{Spuriousness-Guided Training Data Relabeling }\label{sec:training-signals}
% Generating  training labels  spuriousness fine-grained labels
Applying our spuriousness score (Eq. \eqref{eq:spuriousness-score})  to all the possible correlations between the classes and the detected attributes, we can holistically characterize the \textit{spuriousness of images} by showing how likely correlations relevant to an image are exploited by the classifier in predicting the image's label. To analyze the spuriousness of all the training images and pinpoint the robustness pitfalls of the classifier, we propose \textit{spuriousness embedding}, which is defined as follows.

\noindent\textbf{Definition 2 (Spuriousness embedding):}
Given an image $x$ with label $y$, a classifier $f_{\theta}$, the detected attribute set $\mathcal{A}$ with $N_A$ attributes,  a VLM $\phi$, an attribute extraction procedure $\psi$, and the training set $\mathcal{D}_{tr}$, we design the spuriousness embedding for $(x,y)$, denoted as $SE(x,y)$, as a $N_A$-dimensional vector, whose $i_a$-th element is defined as follows,
\begin{equation}\label{eq:spuriousness-embedding}
    SE(x,y)[i_a] = \gamma(a,y;f_\theta,\mathcal{D}_{tr})\cdot \mathds{1}_{{a\in \psi(\phi(x))}}, a\in\mathcal{A},
\end{equation}
where $a$ denotes an attribute of $x$,  $i_a$ denotes the dimension index of $SE(x,y)$ that corresponds to $a$, and $\mathds{1}_{\{\cdot\}}$ is the indicator function that equals one if the condition in the subscript is true and equals zero otherwise.

 With spuriousness embeddings, we embed all images in $\mathcal{D}_{tr}$ in the spuriousness embedding space. Each point in the space represents both an image and a vector characterizing an individual spurious prediction behavior of $f_{\theta}$ in using relevant class-attribute correlations to predict the image. 
 
 In the ideal scenario, a robust classifier produces all zero vectors in the space; practically, we expect to observe a dispersed distribution of points as the classifier does not excessively rely on specific spurious prediction behaviors. In contrast, a biased classifier tends to produce an uneven distribution of points from the same class. We demonstrate in Figure \ref{fig:method-overview}(c) with two clusters of similar prediction behaviors frequently used in predictions, i.e., using the water or land attributes. The partition based on the spuriousness of the images separates same-class samples unevenly, exposing potential prediction failure modes of the classifier, such as predicting a waterbird image with a land background as a landbird.
 
 %while the remaining few points are distant from other same-class points, e.g., images of waterbird with land backgrounds are distant from other waterbird images with water backgrounds. 

Therefore, we cluster images into $K$ clusters in the spuriousness embedding space across different classes to capture potential failure modes in classification.
 Here, $K$ is a hyperparameter and is controlled by our design choice. Formally, we represent the above process as follows,
\begin{equation}\label{eq:clustering}
    p_K(x,y) = CLU(SE(x,y); \mathcal{D}_{tr}, K), \forall(x,y)\in\mathcal{D}_{tr},
\end{equation}
where $p_K(x,y)$ is the cluster label for $(x,y)$, and $CLU$ denotes a clustering algorithm. In this paper, we use KMeans as $CLU$. Now, $p_K(x,y)$ summarizes similar prediction behaviors that may be used by the classifier for predictions. For example, the blue cluster in Figure \ref{fig:method-overview}(c) represents using the general attribute \texttt{water} for predicting waterbird, including using specific attributes such as \texttt{pond} and \texttt{river}.

We use the cluster labels $p_K(x,y)$ to guide the debiasing of the classifier by relabeling the training data with fine-grained labels formulated with $p_K(x,y)$ and $y$, which we unify as one symbol $g_K(x,y) = p_K(x,y)+ (y-1)\cdot K$ ($y$ starts with 1).

\subsection{Learning beyond Classes %: Diversifying Classifier and Balancing Training Data
}\label{sec:training}

To effectively use the relabeled training data for learning robust classifiers, we propose two novel strategies along with training and inference procedures in the following.

\subsubsection{Diversifying Outputs of the Classifier} 
The constructed training labels instruct the classifier that multiple prediction behaviors for the same class should lead to a correct and consistent prediction outcome. To achieve this, we diversify the outputs of the classifier from predicting class labels to distinguishing between different prediction behaviors for the same class.
In this way, the classifier is not only aware of other attributes leading to the same class, but also encouraged to discover more robust features for predictions.

  Specifically, we separate $f_{\theta}$ into a backbone $e_{\theta_1}$ and a $C$-way classification head $q_{\theta_2}$, i.e, $f_{\theta}=q_{\theta_2}\circ e_{\theta_1}$, where $C$ is the number of classes, and the parameters $\theta=\theta_1 \cup \theta_2$. Then, we replace $q_{\theta_2}$ with $h_{\theta_3}$ which is a $(K\cdot C)$-way classification head, resulting in a transformed model $\tilde{f}_{\tilde{\theta}}=h_{\theta_3}\circ e_{\theta_1}$ with $\tilde{\theta}=\theta_3\cup\theta_1$.
Each output of $\tilde{f}_{\tilde{\theta}}$  corresponds to a combination of class $c$ and cluster label $k$. Figure \ref{fig:method-overview}(d) gives an example with $K=2$ and $C=2$.
In this example, instead of predicting two classes, i.e., waterbird and landbird, we replace the classification head of the ERM model with a new classification head to predict the correlations: $\langle$waterbird, \texttt{water}$\rangle$, $\langle$waterbird, \texttt{land}$\rangle$, $\langle$landbird, \texttt{water}$\rangle$, and $\langle$landbird, \texttt{land}$\rangle$.

\subsubsection{Balancing Training Data} 
As discussed in Section \ref{sec:training-signals}, different prediction behaviors may correspond to uneven numbers of samples, which may bias the predictions of the adapted classifier. To address this, we consider within-class and cross-class balancing strategies.
 
\noindent\textbf{Within-class balancing.} We sample $K$ equal-sized training sets $\mathcal{B}_c^k\subseteq \mathcal{G}_c^k$ with images from class $c$ and $K$ different clusters, where $\mathcal{G}_c^k$ is defined as
\begin{equation}
    \mathcal{G}_c^k = \{(x,y)|y=c, p_K(x,y)=k,\forall(x,y)\in\mathcal{D}_{tr}\}.
\end{equation}
In this way, we assign equal importance to predicting the $K$ clusters within the class $c$. We ignore $\mathcal{G}_c^k$ when it is empty.

\noindent\textbf{Cross-class balancing.} Considering that predictions for different classes may show varied reliance on spurious correlations, we additionally balance the size of $\mathcal{B}_c=\cup_{k=1}^K\mathcal{B}_c^k$. Specifically, for each class $c$, we calculate the variance of cluster sizes within class $c$, i.e., $\sigma_c^2=Var(\{|\mathcal{G}_c^k|\ |k=1,\ldots,K\})$, where $|\cdot|$ denotes the size of a set, and $Var(\cdot)$ denotes the variance of a set of numbers. The variance measures the degree of imbalanced prediction behaviors for class $c$. We control the size of $\mathcal{B}_c$ based on $\sigma_c$ such that we sample more training data for a class that exhibits a larger degree of imbalanced prediction behaviors. Concretely, given the batch size $B$, we set $|\mathcal{B}_c|=B\cdot \rho_c$, $|\mathcal{B}_c^k|=B\cdot \rho_c/K$, where $\rho_c=\log(\sigma_c+1)/\sum_{c'=1}^C\log(\sigma_{c'}+1)$, and we use $\log(\sigma_c+1)$ to avoid encountering extreme values.

\subsubsection{Training and Inference} 
The overall learning objective is as follows,
%\vspace{-1mm}
\begin{equation}\label{eq:objective} 
    \tilde{\theta}^* = \arg\min_{\tilde{\theta}} \mathbb{E}_{\mathcal{B}\sim\mathcal{D}_{tr}}\mathbb{E}_{(x,y)\in\mathcal{B}}\ell(\tilde{f}_{\tilde{\theta}}(x),g_K(x,y)),
\end{equation}
%\vspace{-1mm}
where $\mathcal{B}=\cup_{c=1}^C\mathcal{B}_c$.

It is possible that after training on Eq. \eqref{eq:objective}, the model develops reliance on other spurious correlations. Therefore, we iteratively update the spuriousness scores based on the updated model and perform the above mitigation procedure again. We call this method \textit{Learning beyond Classes (LBC)}. The whole procedure is listed in Algorithm 1 in Appendix.

\noindent\textbf{Model selection.}
Given a validation set $\mathcal{D}_{val}$ without group labels,
 we develop a selection metric called \textit{pseudo unbiased validation accuracy} $Acc_{unbiased}^{pseudo}$ to select the best model during training. Specifically, we group the validation data based on the existence of the detected attribute $a$, i.e.,
\begin{equation}
    \mathcal{D}_{val}^a = \{(x,y)|(x,y)\in\mathcal{D}_{val}, a\in\psi(\phi(x))\}.
\end{equation}
Then, we calculate  $Acc_{unbiased}^{pseudo}$ as the average over the accuracy on $\mathcal{D}_{val}^a$ as follows,
\begin{equation}
   Acc_{unbiased}^{pseudo}= \frac{1}{|\mathcal{A}|}\sum_{a\in\mathcal{A}}M(\mathcal{D}_{val}^a;\tilde{f}_{\tilde{\theta}})
\end{equation}

During inference, the predicted label is calculated as $\hat{c} = \lceil(\arg\max_c \tilde{f}_{\tilde{\theta}}(x)) / K\rceil$, where $\lceil c\rceil$ denotes taking the smallest integer greater than or equal to $c$.

\begin{table*}[t]
\centering
\resizebox{\linewidth}{!}{%
\begin{tabular}{llccllllll}
\toprule
\multicolumn{1}{c}{\multirow{2}{*}{Method}}    & \multicolumn{1}{c}{\multirow{2}{*}{Backbone}} & \multicolumn{2}{c}{Group information}                                                                                     & \multicolumn{3}{c}{CelebA}                                                        & \multicolumn{3}{c}{Waterbirds}                                                    \\ \cmidrule(lr){3-10}
\multicolumn{1}{c}{}                           & \multicolumn{1}{c}{}                          & Train                                                       & Validation                                                  & \multicolumn{1}{c}{Worst$(\uparrow)$} & \multicolumn{1}{c}{Average$(\uparrow)$} & \multicolumn{1}{c}{Gap$(\downarrow)$} & \multicolumn{1}{c}{Worst$(\uparrow)$} & \multicolumn{1}{c}{Average$(\uparrow)$} & \multicolumn{1}{c}{Gap$(\downarrow)$} \\ \midrule
GroupDRO \cite{sagawa2019distributionally}        & ResNet-50                                     & \textcolor{red}{\cmark}   & \textcolor{red}{\cmark} & $88.9$                      & $92.9$                        & $4.0$                     & $91.4$                      & $93.5$                        & $2.1$                     \\ \midrule
LfF \cite{nam2020learning}   & ResNet-50                                     & \textcolor{green}{\xmark} & \textcolor{red}{\cmark} & $78.0$                      & $85.1$                        & $7.1$                     & $78.0$                      & $91.2$                        & $13.2$                    \\
 CVaR DRO \cite{levy2020large}       & ResNet-50                                     & \textcolor{green}{\xmark}   & \textcolor{red}{\cmark} & $64.4$                      & $82.5$                        & $18.1$                    & $75.9$                      & $96.0$                       & $20.1$                    \\ 
 JTT \cite{liu2021just} & ResNet-50                                     & \textcolor{green}{\xmark}   & \textcolor{red}{\cmark} & $81.1$                      & $88.0$                        & $6.9$                     & $86.7$                      & $93.3$                        & $6.6$                     \\
 DFR \cite{kirichenko2022last} & ResNet-50                                     & \textcolor{green}{\xmark}   & \textcolor{red}{\cmark} & $69.4_{\pm1.4}$                  & $93.3_{\pm0.1}$                    & $23.9_{\pm 1.3}$                     & $80.2_{\pm2.3}$                  & $92.1_{\pm0.6}$                    & $11.9_{\pm2.7}$                     \\
 \cellcolor[HTML]{EFEFEF}\textbf{LBC (Ours)}                                            & \cellcolor[HTML]{EFEFEF}ResNet-50                                     & \cellcolor[HTML]{EFEFEF}\textcolor{green}{\xmark}   & \cellcolor[HTML]{EFEFEF}\textcolor{red}{\cmark} & \cellcolor[HTML]{EFEFEF}$\textbf{87.4}_{\pm1.8}$                  & \cellcolor[HTML]{EFEFEF}$92.4_{\pm0.3}$                    & \cellcolor[HTML]{EFEFEF}$\textbf{5.0}_{\pm2.1}$                    & \cellcolor[HTML]{EFEFEF}$\textbf{88.1}_{\pm1.4} $                 & \cellcolor[HTML]{EFEFEF}$94.1_{\pm0.3}$                    & \cellcolor[HTML]{EFEFEF}$\textbf{6.0}_{\pm1.7}$            \\ \midrule
ERM & ResNet-50                                     & \textcolor{green}{\xmark}   & \textcolor{green}{\xmark}   & $45.7$                 & $95.5$                     &         $49.8$            & $66.4$                      & $90.2$                        &       $23.8$            \\
LfF \cite{nam2020learning} & ResNet-50                                     & \textcolor{green}{\xmark}   & \textcolor{green}{\xmark}   &       $24.4$         &        $85.1$            &   $60.7$   &     $44.1$          &      $91.2$         &       $47.1$                 \\
CVaR DRO \cite{levy2020large} & ResNet-50                                     & \textcolor{green}{\xmark}   & \textcolor{green}{\xmark}   &     $36.1$       &       $82.5$        &         $46.4$            &          $62.0$        &          $95.2$        &     $33.2$               \\
JTT \cite{liu2021just} & ResNet-50              &  \textcolor{green}{\xmark}  & \textcolor{green}{\xmark} &  $40.6$   &   $88.0$   &     $47.4$   &   $62.5$ &  $93.3$  &      $30.8$         \\
DivDis \cite{lee2022diversify} & ResNet-50              &  \textcolor{green}{\xmark}  & \textcolor{green}{\xmark} &    $55.0$  &     $90.8$   &    $35.8$    &   $81.0$   &  $90.7$  &       $9.7$        \\
MaskTune \cite{asgari2022masktune} & ResNet-50              &  \textcolor{green}{\xmark}  & \textcolor{green}{\xmark} &  $78.0_{\pm 1.2}$   &    $91.3_{\pm 0.1}$   &   $13.3_{\pm1.3}$     &   $86.4_{\pm 1.9}$   &   $93.0_{\pm 0.7}$ &       $6.6_{\pm2.6}$        \\
DFR \cite{kirichenko2022last} & ResNet-50              &  \textcolor{green}{\xmark}  & \textcolor{green}{\xmark} & $46.0$   &    $95.8$    &    $49.8$    &  $77.4$   &   $92.1$   &       $14.7$        \\
\cellcolor[HTML]{EFEFEF}\textbf{LBC (Ours)}   & \cellcolor[HTML]{EFEFEF}ResNet-50                                     & \cellcolor[HTML]{EFEFEF}\textcolor{green}{\xmark}   &  \cellcolor[HTML]{EFEFEF}\textcolor{green}{\xmark} & \cellcolor[HTML]{EFEFEF}$\textbf{81.2}_{\pm1.5}$                  & \cellcolor[HTML]{EFEFEF}$92.2_{\pm0.3}$                    & \cellcolor[HTML]{EFEFEF}$\textbf{11.0}_{\pm1.8} $                    & \cellcolor[HTML]{EFEFEF}$\textbf{87.3}_{\pm1.8} $                  & \cellcolor[HTML]{EFEFEF}$93.2_{\pm0.9}$                    & \cellcolor[HTML]{EFEFEF}$\textbf{5.9}_{\pm2.7} $                     \\ 

\bottomrule         
\end{tabular}
}
%\vspace{-2mm}
\caption{Worst-group and average accuracy (\%) comparison with state-of-the-art methods on the CelebA and Waterbirds datasets. The ResNet-50 backbones are pre-trained on ImageNet. GroupDRO reveals the theoretically best performance given all the group information in worst-group results and performance gaps. The best worst-group results and performance gaps are in \textbf{boldface}. 
}\label{tab:celeba-waterbirds}
%\vskip -6pt
\end{table*}

\section{Experiment}\label{sec:experiment}

\subsection{Datasets}

\noindent \textbf{Waterbirds} \cite{sagawa2019distributionally} is a dataset for recognizing waterbirds and landbirds. It is generated synthetically by combining images of the two kinds of birds from the CUB dataset \cite{WelinderEtal2010} and the backgrounds, water and land,  from the Places dataset \cite{zhou2017places}.

\noindent \textbf{CelebA} \cite{liu2015deep} is a large-scale image dataset of celebrity faces. The task is to identify hair color, non-blond or blond, with male and female as the spurious attributes.

\noindent \textbf{ImageNet-9} \cite{xiao2021noise} comprises images with different background and foreground signals, which can be used to assess how much models rely on image backgrounds. This dataset is a subset of ImageNet \cite{imagenet} containing nine super-classes.

\noindent \textbf{ImageNet-A} \cite{hendrycks2021natural} is a dataset of real-world images, adversarially curated to test the limits of classifiers such as ResNet-50. While these images are from standard ImageNet classes \cite{imagenet}, their complexity increases the challenge, often causing misclassifications in multiple models. We use this dataset to test the robustness of a classifier after training it on ImageNet-9.

\noindent \textbf{NICO} \cite{he2021towards} is designed for non-independent and identically distributed and out-of-distribution image classification, simulating real-world scenarios where testing distributions differ from training ones. It labels images with both main concepts and contexts (e.g., 'dog on grass'), enabling studies on transfer learning, domain adaptation, and generalization.

Details of the dataset settings are shown in Appendix.

\subsection{Experimental Setup}
\paragraph{Spurious attribute detection.} We generate text descriptions for images using a pre-trained vision-language model \cite{nlp_connect_2022}, which has an encoder-decoder structure where the encoder is a vision transformer \cite{dosovitskiyimage}  and the decoder is the GPT-2 \cite{radford2019language} language model. We set the maximum length of the sequence to be generated as 16 and the number of beams for beam search to 4. After generating text descriptions for test images, we use Spacy (\url{https://spacy.io/}) to extract nouns and adjectives from the descriptions automatically. We additionally filter out words with frequencies less than 10 to remove potential annotation noise. In our experiments, we only need to do this procedure once for each dataset.

\paragraph{Training settings.} We use ResNet-50 and ResNet-18 as the backbone networks. For each dataset, we first train an ERM model, which is first initialized with ImageNet pre-trained weights, for 100 epochs. We set the learning rate to 0.001 which decays following a cosine annealing scheduler and use an SDG optimizer with 0.9 momentum and $10^{-4}$ weight decay. Then, we use the ERM-trained models as the initial models for our LBC training. Standard data augmentations are used in LBC to effectively mitigate spurious correlations that are not typically captured by VLMs, such as sizes and orientations. For all the datasets, we fix the learning rate to 0.0001 and the batch size to 128. We sample 20 batches per epoch and train for 50 epochs. The cluster size $K$ is set to 3. We report our results averaged over 5 runs. We provide training details in the Appendix. All experiments are conducted on NVIDIA RTX 8000 GPUs \footnote{Our code is available at \url{https://github.com/gtzheng/LBC}.}.

\paragraph{Evaluation metrics.} We adopt different evaluation metrics on different datasets. For a dataset with group labels defined by classes and biased attributes, we partition the test data into groups. \textit{Average accuracy} measures the overall performance of a classifier on the test data; however, it may be dominated by the majority group of samples with certain biases that the classifier may heavily rely on for predictions. Therefore, we mainly focus on the \textit{worst-group accuracy} which is a widely accepted robustness measure that gives the lower-bound performance of a classifier on various dataset biases. To measure the tradeoff between the average and worst accuracy, we additionally calculate the \textit{gap} between the two metrics. For datasets without group labels, we report different kinds of average accuracies on specifically designed test sets, which we will explain in the respective sections.

\begin{table}[thbp]
\small
\begin{tabular}{cL{6cm}}
\toprule
Class     & \multicolumn{1}{c}{Attributes}    \\ \midrule
\multirow{2}{*}{landbird} & \colorbox{blue!10}{pool}, \colorbox{blue!10}{boat}, \colorbox{orange!10}{building}, \colorbox{blue!10}{pond}, \colorbox{blue!10}{surfboard}, \colorbox{orange!10}{sandy}, \colorbox{blue!10}{beach}, \colorbox{blue!10}{water}, body, frisbee       \\ \midrule
\multirow{2}{*}{waterbird} & \colorbox{blue!10}{stream}, \colorbox{orange!10}{forest}, \colorbox{orange!10}{building}, pile, front, middle, animal, photo, \colorbox{orange!10}{tree}, branch\\ \bottomrule
\end{tabular}
% \vskip -4pt
\caption{Top-10 detected attributes selected based on their spuriousness scores for each class in the Waterbirds dataset. We highlight several attributes that are relevant to water backgrounds in blue and those that are relevant to land backgrounds in orange.}\label{tab:top-attributes}
% \vskip -12pt
\end{table}

\begin{table*}[!htb]
    \hspace{-0.5cm}
    \begin{minipage}{.55\linewidth}
      \centering
      \resizebox{1.055\linewidth}{!}{
        \begin{tabular}{lclll}
        \toprule
        \multicolumn{1}{c}{\multirow{2}{*}{Method}} & \multirow{2}{*}{\begin{tabular}[c]{@{}c@{}}Spurious\\ attribute label\end{tabular}} & \multicolumn{2}{c}{ImageNet-9}                                & \multicolumn{1}{c}{ImageNet-A} \\ \cmidrule{3-5} 
        \multicolumn{1}{c}{}                        &                                                                                     & \multicolumn{1}{c}{Validation$(\uparrow)$} & \multicolumn{1}{c}{Unbiased$(\uparrow)$} & \multicolumn{1}{c}{Test$(\uparrow)$}       \\ \midrule
        StylisedIN \cite{geirhos2018imagenet}                                                                      & \textcolor{red}{\cmark}                                                                                   & $88.4_{\pm0.5}$                       & $86.6_{\pm0.6}$                     & $24.6_{\pm1.4}$                       \\
        LearnedMixin \cite{clark2019don}                                                                   & \textcolor{red}{\cmark}                                                                                   & $64.1_{\pm4.0}$                       & $62.7_{\pm3.1}$                     & $15.0_{\pm1.6}$                       \\
        RUBi \cite{cadene2019rubi}                                                                           & \textcolor{red}{\cmark}                                                                                   & $90.5_{\pm0.3}$                      & $88.6_{\pm0.4}$                     & $27.7_{\pm2.1}$                       \\ \midrule
        ERM                                                                             & \textcolor{green}{\xmark}                                                                                   & $90.8_{\pm0.6}$                       & $88.8_{\pm0.6}$                     & $24.9_{\pm1.1}$                       \\
        % Biased (BagNet18)  \cite{brendel2019approximating}                                                              & \textcolor{green}{\xmark}                                                                                   & 67.7±0.3                       & 65.9±0.3                     & 18.8±1.15                      \\
        ReBias \cite{bahng2020learning}                                                                         & \textcolor{green}{\xmark}                                                                                   & $91.9_{\pm1.7}$                       & $90.5_{\pm1.7}$                     & $29.6_{\pm1.6}$                       \\
        LfF \cite{nam2020learning}                                                                             & \textcolor{green}{\xmark}                                                                                   & $86.0$                             & $85.0$                           & $24.6$                           \\
        CaaM \cite{wang2021causal}                                                                            & \textcolor{green}{\xmark}                                                                                   & $95.7$                           & $95.2$                         & $32.8$                           \\
        SSL+ERM \cite{kim2022learning}                                                                         & \textcolor{green}{\xmark}                                                                                   & $94.18_{\pm0.07}$                     & $93.18_{\pm0.04}$                   & $34.21_{\pm0.49}$                     \\
        LWBC \cite{kim2022learning}                                                                            & \textcolor{green}{\xmark}                                                                                   & $94.03_{\pm0.23}$                    & $93.04_{\pm0.32}$                   & $35.97_{\pm0.49}$                     \\
        \cellcolor[HTML]{EFEFEF}\textbf{LBC (Ours)}                                                                            &\cellcolor[HTML]{EFEFEF} \textcolor{green}{\xmark}                                                                                   &\cellcolor[HTML]{EFEFEF}$\textbf{96.97}_{\pm0.17}$                     & \cellcolor[HTML]{EFEFEF}$\textbf{96.03}_{\pm0.12}$                   &\cellcolor[HTML]{EFEFEF}$\textbf{40.63}_{\pm1.79}$                     \\ \bottomrule
        \end{tabular}
        }
        % \vskip -4pt
    \caption{Validation, Unbiased, and Test metrics (\%) evaluated on the ImageNet-9 and ImageNet-A datasets. All methods use ResNet-18 as the backbone. The best results are in \textbf{boldface}.}
    \label{tab:imagenet}
    \end{minipage}%
    \hspace{0.5cm}
    \begin{minipage}{.457\linewidth}
      \centering
        \resizebox{\linewidth}{!}{
        \begin{tabular}{lcll}
        \toprule
        \multicolumn{1}{c}{}                         &                                                                                      & \multicolumn{2}{c}{NICO}                                  \\ \cmidrule{3-4} 
        \multicolumn{1}{c}{\multirow{-2}{*}{Method}} & \multirow{-2}{*}{\begin{tabular}[c]{@{}c@{}}Spurious\\ attribute label\end{tabular}} & \multicolumn{1}{c}{Validation$(\uparrow)$} & \multicolumn{1}{c}{Test$(\uparrow)$} \\ \midrule
        % Cutout                                                                           & \textcolor{red}{\cmark}                                                                                   & 43.69                          & 43.77                    \\
        RUBi \cite{cadene2019rubi}                                                                            & \textcolor{red}{\cmark}                                                                                   & $43.86$                          & $44.37$                    \\
        IRM \cite{arjovsky2019invariant}                                                                             & \textcolor{red}{\cmark}                                                                                   & $40.62$                          & $41.46$                    \\
        % Unshuffle                                                                        & \textcolor{red}{\cmark}                                                                                   & 43.15                          & 43                       \\
        % REx                                                                              & \textcolor{red}{\cmark}                                                                                   & 41                             & 41.15                    \\ 
        \midrule
        ERM                                                                              & \textcolor{green}{\xmark}                                                                                   & $43.77$                          & $42.61$                    \\
        CBAM  \cite{woo2018cbam}                                                                           & \textcolor{green}{\xmark}                                                                                   & $42.15$                          & $42.46$                    \\
        ReBias \cite{bahng2020learning}                                                                          & \textcolor{green}{\xmark}                                                                                   & $44.92$                          & $45.23$                   \\
        LfF \cite{nam2020learning}                                                                             & \textcolor{green}{\xmark}                                                                                   & $41.83$                          & $40.18$                    \\
        CaaM  \cite{wang2021causal}                                                                         & \textcolor{green}{\xmark}                                                                                   & $46.38$                          & $46.62$                    \\
        SSL+ERM \cite{kim2022learning}                                                                         & \textcolor{green}{\xmark}                                                                                   & $55.63_{\pm0.54}$                     & $52.24_{\pm0.27}$               \\
        LWBC \cite{kim2022learning}                                                                            & \textcolor{green}{\xmark}                                                                                   & $56.05_{\pm0.45}$                     & $52.84_{\pm0.31}$              \\
        \cellcolor[HTML]{EFEFEF}\textbf{LBC (Ours)}                                                                          & \cellcolor[HTML]{EFEFEF}\textcolor{green}{\xmark}                                                                                   & \cellcolor[HTML]{EFEFEF}$\textbf{68.26}_{\pm2.15}$                     & \cellcolor[HTML]{EFEFEF}$\textbf{65.34}_{\pm2.54}$               \\ \bottomrule
        \end{tabular}
        }
        % \vskip -4pt
        \caption{Validation and Test metrics (\%) evaluated on the NICO dataset. All methods use ResNet-18 as the backbone pre-trained on ImageNet. The best results are in \textbf{boldface}.}
        \label{tab:nico}
    \end{minipage} 
% \vskip -10pt
\end{table*}

\subsection{Effectiveness of Spuriousness Score}
We calculated the spuriousness scores for the correlations between all the detected attributes and classes in the Waterbirds dataset and selected the top-10 attributes with the highest spuriousness scores for each class. 
Table \ref{tab:top-attributes} shows that these attributes are mostly relevant to water and land backgrounds, which are spurious by design. Interestingly, our spuriousness score can find attributes in one class that are heavily exploited to predict some other class. For example, \texttt{pool} from landbird is detected in images of landbird with a pool, but it tends to bias the predictions toward waterbird since it is relevant to water backgrounds. As discussed in Section \ref{sec:spurious-detection}, this arises from the Abs operator in our definition of spuriousness score in Eq. \eqref{eq:spuriousness-score}: when the nominator term is much smaller than the denominator term, e.g., when $a$=\texttt{pool} and $c$=landbird, the spuriousness score is still high. Examples from other datasets are shown in the Appendix.

\subsection{LBC Reduces Reliance on Spurious Correlations}
\begin{figure}[thbp]
    \centering
    % \vskip -6pt
    \includegraphics[width=1.0\linewidth]{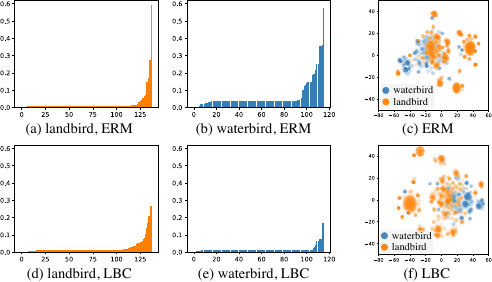}
    % \vskip -2pt
    \caption{(a) and (b): Spuriousness scores for the attributes detected from landbird and waterbird based on an ERM model. (d) and (e): Spuriousness scores based on our LBC model. (c) and (f): Spuriousness embeddings of the images in the Waterbirds dataset based on the ERM and LBC model, respectively.}
    \label{fig:effectiveness-lbc}
     % \vskip -6pt
\end{figure}

 To show the efficacy of our LBC method in learning a model robust to spurious correlations, we quantitatively and qualitatively demonstrate this by using spuriousness scores and visualizations of spuriousness embeddings, respectively.
We first calculated the spuriousness scores for the class-attribute correlations based on an ERM and our LBC-trained models. Comparing the results between Figure \ref{fig:effectiveness-lbc}(a) and Figure \ref{fig:effectiveness-lbc}(d), as well as between Figure \ref{fig:effectiveness-lbc}(b) and Figure \ref{fig:effectiveness-lbc}(e), we observe that LBC significantly reduces the spuriousness scores of the correlations between the detected attributes and the two classes. Moreover, the spuriousness embeddings, which represent images with different prediction behaviors, become more dispersed in Figure \ref{fig:effectiveness-lbc}(f) than in Figure \ref{fig:effectiveness-lbc}(c). This indicates that LBC successfully mitigates the ERM classifier's reliance on certain prediction behaviors and diversifies the prediction behaviors for different classes.

\subsection{Comparison with Existing Methods}
\textbf{Datasets with group labels.} Table \ref{tab:celeba-waterbirds} gives the results on the CelebA and Waterbirds datasets in two settings. In the first setting where only the group labels of the validation data are used, our method achieves the best worst-group accuracies and the best gaps between the average and the worst-group accuracies on the two datasets, striking a favorable balance between the robustness of the classifier and its overall performance. Our worst-group accuracies are also close to the upper bounds established by GroupDRO, while our average accuracies are competitive or better than those of GroupDRO. In our main setting where no group labels are available, our method
outperforms the baselines with the best worst-group accuracies and worst-average gaps. More significantly, our method is most robust in terms of the least drops in worst-group accuracy when switching from the first setting to the second one. This shows the effectiveness of our designed model selection metric %(Section \ref{sec:model-selection}) 
in selecting robust models.

\noindent\textbf{Datasets with texture biases.} Experiments on ImageNet-9 and ImageNet-A test how much a classifier relies on the spurious texture bias. In Table \ref{tab:imagenet}, texture group labels \cite{bahng2020learning,kim2022learning} are used as the spurious attribute labels. The ``Validation" denotes the average accuracy on the validation set,  ``Unbiased" denotes the average accuracy over several texture groups, and ``Test" denotes the average accuracy on the ImageNet-A dataset which contains misclassified samples by an ImageNet-trained model. Our method outperforms other methods on the three metrics, showing that our method is effective in mitigating a classifier's reliance on texture biases and correcting its failure modes in classification.

\noindent\textbf{Dataset with object-context correlations.} The NICO dataset is created to evaluate a classifier's reliance on object-context correlations.  Images in the training set are long-tailed distributed in the sense that an object class has exponentially decreasing numbers of images that correlate with the contexts. In Table \ref{tab:nico}, ``Validation" denotes the average accuracy on the validation set, and ``Test" denotes the average accuracy on the test set. Both validation and test sets  contain not only existing object-context correlations but also unseen ones. Our method effectively mitigates the reliance on object-context correlations and achieves the best on the two metrics.

\subsection{Ablation Study}
\begin{figure}[thbp]
    \centering
    % \vskip -4pt
    \includegraphics[width=0.8\linewidth]{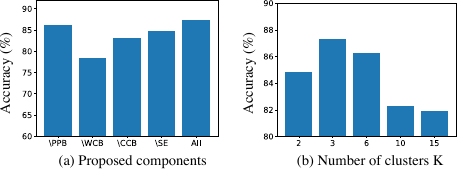}
    % \vskip -4pt
    \caption{Worst-group accuracy comparison of (a) leave-one-out study on the four proposed components and (b) analysis on the number of clusters $K$ on the Waterbirds dataset.}
    \label{fig:ablation-results}
    % \vspace{-0.3in}
\end{figure}

We first analyzed the effectiveness of the four proposed components: (1) predicting prediction behaviors (PPB), (2) within-class balancing (WCB), (3) cross-class balancing, and (4) spuriousness embeddings (SE).  We modify one component and observe the worst-group accuracies achieved by the remaining ones. In Figure \ref{fig:ablation-results}(a), $\backslash$PPB denotes that we keep the original classifier to predict classes, $\backslash$WCB denotes that we randomly sample from the same class, $\backslash$CCB denotes that we equally sample images with different identified prediction behaviors, and $\backslash$SE denotes that we use binary attribute embeddings (i.e., $SE(x,y)[i_a] = \mathds{1}_{{a\in \psi(\phi(x))}}$ in Eq. \eqref{eq:spuriousness-embedding}) for images. We observe that all four components positively contribute to our method since replacing any one of them results in reduced accuracy. Among the four components,  balanced data sampling (WCB and CCB), especially WCB, is most critical to our method.
Figure \ref{fig:ablation-results}(b) shows that a large $K$ exceeding 10 has suboptimal worst-group accuracies and that when $K=2$, it limits the discovery of diverse prediction behaviors. Typically, $K=3$ works for most of the cases. We also showed in Appendix that the detected attributes from the VLM alone do not contain information effective for improving a classifier's robustness to spurious correlations.

% \vspace{-0.3in}
\section{Conclusion}
We completely removed the barrier of requiring expert knowledge and human annotations for spurious correlation mitigation by proposing a self-guided framework. Our framework incorporates an automatic approach empowered by a VLM to detect attributes in images and quantifies their spuriousness with class labels. We formulated a spuriousness embedding space based on the spuriousness scores to identify distinct prediction behaviors of a classifier. We trained the classifier to recognize the identified prediction behaviors with balanced training data. Experiments showed that our framework improves the  robustness of a classifier against spurious correlations without knowing them in the data. 
 \appendix
 \section*{Acknowledgments}
This work is supported in part by the US National Science Foundation under grants 2217071, 2213700, 2106913, 2008208, 1955151.
\bibliographystyle{named}
\bibliography{references}

\clearpage
% \renewcommand{\thetable}{A\arabic{table}}
% \renewcommand{\thefigure}{A\arabic{figure}}
% \renewcommand{\theequation}{A\arabic{equation}}
% \setcounter{table}{0}
% % \setcounter{page}{0}
% \setcounter{equation}{0}
% \setcounter{theorem}{0}
% % \renewcommand{\thesubsection}{\Alph{subsection}}
% \clearpage
\appendix
\noindent {\Large\bf Appendix}
\section{Learning Algorithm}
The whole learning procedure of our proposed LBC is shown in Algorithm \ref{alg:1}.  We iteratively retrain a model adapted from an ERM-trained model using relabeled (Section \ref{sec:training-signals}) and balanced (Section \ref{sec:training}) training data. Our relabeling does not alter the class membership of the training data; instead, it creates fine-grained labels within classes. Therefore, although the classification head of $\tilde{\theta}$ keeps changing in each training epoch, the model's ability to recognize different classes keeps improving after each training epoch.
Even when the generated fine-grained labels are noisy, the backbone of $\tilde{\theta}$ is still encouraged to recognize different classes. We select the best model based on its performance on the validation data.

\noindent\textbf{Time complexity.} The first step of our algorithm, i.e., building the attribute set $\mathcal{A}$, is only needed once for each dataset. Thus, its time complexity is negligible once $\mathcal{A}$ has been generated. Generating the spurious scores needs a forward pass of all the $N$ training samples. Therefore, the time complexity is $O(N)$ with a scaling constant $\tau_{sc}$ representing the average complexity over $N$ samples. The KMeans clustering step has a time complexity of $O(KNT)$ with a scaling constant $\tau_{clu}$, where $K$ is the number of clusters, $T$ is the number of iterations for the clustering, and $\tau_{clu}$ denotes the complexity for computing the Euclidean distance between two vectors. The complexity of optimizing $\tilde{f}_{\tilde{\theta}}$ is $O(N)$ with a scaling constant $\tau_{opt}$ denoting the complexity for a backward pass of the model. Typically, $\tau_{sc}\ll \tau_{opt}$, $\tau_{clu}\ll \tau_{opt}$, and $K\cdot T$ is typicall small. Therefore, the overall complexity of our algorithm is approximately $O(EN)$, where $E$ is the number of training epochs.
%, and $\tau_{sc}$ denotes the time required for a forward pass of the model.
%and $\tau_{clu}$ denotes the complexity for computing the Euclidean distance between two vectors. 

\begin{algorithm}[h]
		\caption{Learning beyond classes (LBC)}
		$\mathbf{Input}$: Training dataset $\mathcal{D}_{tr}$, an ERM trained model $f_{\theta}$, number of clusters $K$, a pre-trained VLM $\phi$, an attribute extraction procedure $\psi$, and the number of training iterations $E$.\\
		$\mathbf{Output}$: Learned weights $\tilde{\theta}$
		\begin{algorithmic}[1]
        \STATE{Build the attribute set $\mathcal{A}=\cup_{(x,y)\in\mathcal{D}_{tr}}\psi(\phi(x))$}
        \STATE{Transform $f_{\theta}$ into $\tilde{f}_{\tilde{\theta}}$}
        \FOR{$e=1,\ldots,E$}
		\STATE Generate spuriousness scores using Eq. \eqref{eq:spuriousness-score}
        \STATE{Get cluster labels $p_K(x,y)$ with Eq. \eqref{eq:clustering}}
        \STATE{Relabeling with $g_K(x,y) = p_K(x,y)+ (y-1)\cdot K$}
        \STATE{Optimize $\tilde{f}_{\tilde{\theta}}$ using Eq. \eqref{eq:objective}}
        \ENDFOR
        \RETURN $\tilde{\theta}$ 
		\end{algorithmic}\label{alg:1}
\end{algorithm}
\section{Datasets}
Table \ref{tab:dataset-statistics} depicts detailed statistics for all datasets.  For Waterbirds and CelebA datasets, we give the number of training, validation, and test images in each group specified by classes and attributes. For example, the group $\langle$landbird, land$\rangle$ in the Waterbirds dataset has 3498 training images which are all landbirds and have land backgrounds. The NICO dataset uses multiple contexts as spurious attributes which are listed in Table \ref{tab:class-context-nico}.  The ImageNet-9 and ImageNet-A datasets do not have clear group partitions specified by the class and attribute associations. 

\begin{table}[htbp]
\resizebox{\linewidth}{!}{%
\begin{tabular}{lccccc}
\hline
\multicolumn{1}{l}{\multirow{2}{*}{Dataset}} & \multicolumn{1}{l}{\multirow{2}{*}{\begin{tabular}[c]{@{}c@{}}Number of \\ classes\end{tabular}}} & \multicolumn{1}{c}{\multirow{2}{*}{$\langle$class, attribute$\rangle$}} & \multicolumn{3}{c}{Number of images} \\ \cline{4-6} 
\multicolumn{1}{l}{}                         & \multicolumn{1}{l}{}                                   & \multicolumn{1}{l}{}                                                  & Train       & Val        & Test      \\ \hline
\multirow{4}{*}{Waterbirds}                  & \multirow{4}{*}{2}                                     & $\langle$landbird, land$\rangle$                                        & 3,498       & 467        & 2,255     \\
                                             &                                                        & $\langle$landbird, water$\rangle$                                       & 184         & 466        & 2,255     \\
                                             &                                                        & $\langle$waterbird, land$\rangle$                                       & 56          & 133        & 642       \\
                                             &                                                        & $\langle$waterbird, water$\rangle$                                      & 1,057       & 133        & 642       \\ \hline
\multirow{4}{*}{CelebA}                      & \multirow{4}{*}{2}                                     & $\langle$non-blond, female$\rangle$                                     & 71,629      & 8,535      & 9,767     \\
                                             &                                                        & $\langle$non-blond, male$\rangle$                                       & 66,874      & 8,276      & 7,535     \\
                                             &                                                        & $\langle$blond, female$\rangle$                                         & 22,880      & 2,874      & 2,480     \\
                                             &                                                        & $\langle$blond, male$\rangle$                                           & 1,387       & 182        & 180       \\ \hline
NICO                                         & 10                                                     & -                                                                     & 2840        & 1299       & 1299      \\ \hline
ImageNet-9                                   & 9                                                      & -                                                                     & 54,600      & 2,100      & -         \\ \hline
ImageNet-A                                   & 9                                                      & -                                                                     & -           & -          & 1087      \\ \hline
\end{tabular}%
}\caption{Detailed statistics of the 5 datasets. $\langle$class, attribute$\rangle$ represents a spurious correlation between a class and a spurious attribute. ``-" denotes not applicable.}\label{tab:dataset-statistics}
\end{table}

In the NICO dataset \cite{wang2021causal}, the training set consists of 7 context classes per object class, and there are 10 object classes. Images in the training set are long-tailed distributed in the sense that an object class has exponentially decreasing numbers of images that correlate with the 7 contexts. Table \ref{tab:class-context-nico} gives the contexts for each of the 10 classes. The contexts of a class are arranged based on the number of images they have in the class, and the first context has the most images. We follow the setting in \cite{wang2021causal}, where for each class, the number of images having the context $t$ is proportional to the ratio $\text{IR}^{i_t}$, where $i_t (0\leq i_t \leq 6$ denotes the index of the context $t$ for the corresponding class in  Table \ref{tab:class-context-nico}, and the imbalance ratio IR is 0.02. The validation and test sets contain images from the 10 classes, and each class has a equal number of images from its 7 associated context classes and 3 new contexts not seen in the training.

\begin{table}[htbp]
%\resizebox{\linewidth}{!}{%
\small
\begin{tabular}{cL{6.5cm}}
\hline
Class    & \multicolumn{1}{c}{Contexts}                                                                            \\ \hline
dog      & on\_grass,  in\_water, in\_cage, eating, on\_beach, lying, running; at home, in street, on snow         \\
cat      & on\_snow, at\_home, in\_street, walking, in\_river, in\_cage, eating; in water, on grass, on tree       \\
bear     & in\_forest, black, brown, eating\_grass, in\_water, lying, on\_snow; on ground, on tree, white          \\
bird     & on\_ground, in\_hand, on\_branch, flying, eating, on\_grass, standing; in water, in cage, on shoulder   \\
cow      & in\_river, lying, standing, eating, in\_forest, on\_grass, on\_snow; at home, aside people, spotted     \\
elephant & in\_zoo, in\_circus, in\_forest, in\_river, eating, standing, on\_grass; in street, lying, on snow      \\
horse    & on\_beach, aside\_people, running, lying, on\_grass, on\_snow, in\_forest; at home, in river, in street \\
monkey   & sitting, walking, in\_water, on\_snow, in\_forest, eating, on\_grass; in cage, on beach, climbing       \\
rat      & at\_home, in\_hole, in\_cage, in\_forest, in\_water, on\_grass, eating; lying, on snow, running         \\
sheep    & eating, on\_road, walking, on\_snow, on\_grass, lying, in\_forest; aside people, in water, at sunset    \\ \hline
\end{tabular}%
%}
\caption{Classes and their associated contexts in the NICO datasets. Contexts after the semicolons are unseen in the training set.}\label{tab:class-context-nico}
\end{table}

The ImageNet-9 dataset \cite{bahng2020learning} is a subset of ImageNet. It has 9 super-classes, i.e., Dog, Cat, Frog, Turtle, Bird, Primate, Fish, Crab, Insect, which are obtained by merging similar classes from ImageNet. ImageNet-A contains real-world images
that are challenging to the image classifiers trained on standard ImageNet.  We extract images of the 9 super-classes from the ImageNet-A dataset and use these images as the test data. To calculate the Unbiased accuracy on the validation set of ImageNet-9, we use the cluster labels provided in \cite{bahng2020learning} that partition the validation data into groups and calculate the average accuracy over these groups.

\section{Implementation Details}
\subsection{Spurious Attribute Detection} 
\begin{figure}[htbp]
    \centering
    \includegraphics[width=\linewidth]{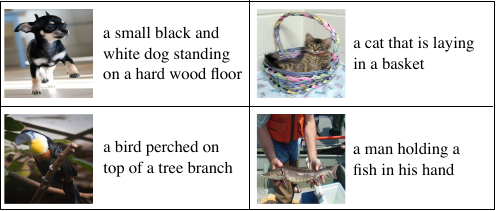}
    \caption{Examples of the generated text descriptions for images in the ImageNet-9 dataset.}
    \label{fig:text-description-examples}
\end{figure}

We generate text descriptions for images using a pre-trained ViT-GPT2 model \cite{nlp_connect_2022}. Fig. \ref{fig:text-description-examples} shows four images from the ImageNet-9 dataset along with their descriptions generated by the ViT-GPT2 model. After generating text descriptions, we use Spacy (https://spacy.io/) to automatically extract nouns and adjectives from the descriptions. Then, we add the extracted words to $\mathcal{A}$, forming a set of detected attributes which are potentially spurious. We additionally filter out attributes with frequencies less than 10 to remove rare words that represent too few images and potential annotation noise. Table \ref{tab:attribute-statistics} shows the numbers of detected attributes as well as the numbers of average detected attributes per image for the four datasets which we used during training.  We did not detect attributes on the ImageNet-A dataset since it was only used for testing.

\begin{table}[htbp]
%\resizebox{\linewidth}{!}{%
\begin{tabular}{ccc}
\hline
Dataset    & \begin{tabular}[c]{@{}c@{}}Number of \\ detected attributes\end{tabular} & \begin{tabular}[c]{@{}c@{}}Average number of \\ attributes per image\end{tabular} \\ \hline
Waterbirds & 144                                                                      & 4.314                                                                             \\
CelebA     & 345                                                                      & 4.291                                                                             \\
NICO       & 199                                                                      & 3.995                                                                             \\
ImageNet-9 & 442                                                                      & 4.311                                                                             \\ \hline
\end{tabular}%
%}
\caption{Statistics of the attributes detected from the Waterbirds, CelebA, NICO, and ImageNet-9 datasets.}\label{tab:attribute-statistics}
\end{table}

\begin{table*}[htbp]
%\resizebox{\linewidth}{!}{%
\centering
\begin{tabular}{ccccccc}
\hline
Dataset    & Backbone  & Initialization       & Learning Rate & \begin{tabular}[c]{@{}c@{}}Learning Rate \\ Scheduler\end{tabular} & Batch Size & Epochs \\ \hline
Waterbirds & ResNet-50 & ImageNet pre-trained & 3e-3          & Cosine Annealing                                                   & 32         & 100    \\
CelebA     & ResNet-50 & ImageNet pre-trained & 3e-3          & Cosine Annealing                                                   & 100        & 20     \\
NICO       & ResNet-18 & ImageNet pre-trained & -          & -                                                  & -        & -    \\
ImageNet-9 & ResNet-18 & ImageNet pre-trained & 1e-3          & Cosine Annealing                                                    & 128        & 100    \\ 
ImageNet-9 & ResNet-18 & Random & 5e-2          & MultiStepLR([50, 80, 100],0.2)                                                     & 256        & 100    \\ \hline
\end{tabular}%
%}
\caption{Details for training ERM models on the four datasets. MultiStepLR([epoch1, epoch2, epoch3], $r$) denotes a learning rate scheduler which decays the learning rate at specified epochs with a multiplication factor $r$, and `-' denotes no training.}\label{tab:emr-training-details}
\end{table*}

\begin{table*}[htbp]
\centering
%\resizebox{\linewidth}{!}{%
\begin{tabular}{ccccccc}
\hline
Dataset    & Learning rate & Batch size & \begin{tabular}[c]{@{}c@{}}Number of Batches \\ Per Epoch\end{tabular} & Training Epochs & K & \begin{tabular}[c]{@{}c@{}}Model Selection\\ Metric\end{tabular}               \\ \hline
Waterbirds & 1e-4          & 128        & 20                                                                     & 50     & 3 & PU-ValAcc \\
CelebA     & 1e-4          & 128        & 20                                                                     & 50     & 3 & PU-ValAcc \\
NICO       & 1e-4          & 128        & 50                                                                     & 50     & 3 & PU-ValAcc                                                           \\
ImageNet-9 & 1e-4          & 128        & 200                                                                     & 100     & 4 & Validation accuracy                                                            \\ \hline
\end{tabular}%
%}
\caption{Hyperparameter settings and model selection criteria for LBC training on the Waterbirds, CelebA, NICO, and ImageNet-9 datasets. PU-ValAcc denotes pseudo unbiased validation accuracy.}\label{tab:lbc-training-details}
\end{table*}

\begin{figure}[thbp]
    \centering
    \includegraphics[width=\linewidth]{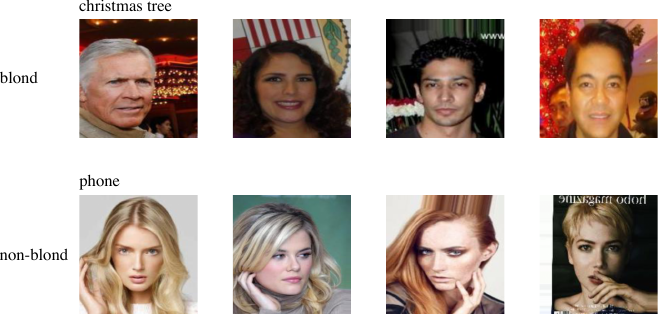}
    \caption{Samples selected based on the two detected attributes, \texttt{christmas tree} and \texttt{phone}. Although these attributes are not self-explanatory in representing the selected samples, samples selected by them have some common characteristics.}
    \label{fig:task-examples}
\end{figure}

\paragraph{Non-self-explanatory attributes are still informative.}
We use two detected attributes, \texttt{christmas tree} and \texttt{phone}, to select samples from the CelebA dataset and show four samples for each of the attribute in Fig. \ref{fig:task-examples}. We observe that \texttt{christmas tree} and \texttt{phone} are not self-explanatory in representing the common features shared among the samples because of the limited capacity of the pre-trained vision-language model (VLM) used to generate text descriptions. However,  samples selected by each attribute do have some characteristics shared in common. For the samples  selected based on \texttt{christmas tree}, they all have background colors that are related to a Christmas tree, e.g., red colors are recognized as some decorations on a Christmas tree by the pre-trained VLM. In the samples selected based on \texttt{phone}, the people all hold their hand close to their faces.

\subsection{Training ERM Models}
Our method starts with an ERM model and retrains it in an adapted form using the proposed techniques. Table \ref{tab:emr-training-details} shows the detailed settings for training ERM models on the four datasets. Note that for the NICO dataset, since the training data is limited, we did not use the training data to train an ERM model; instead, we followed the setting in \cite{kim2022learning} to only initialize a model for the later LBC training with ImageNet pre-trained weights. For fair comparison with existing methods, we adopted ResNet-50 as the backbone for experiments on the Waterbirds and CelebA datasets and adopted ResNet-18 as the backbone for experiments on the NICO and ImageNet-9 datasets. All images are resized to $224\times224$ resolution. Standard data augmentations, i.e., RandomResizedCrop and RandomHorizontalFlip, were used in training these models. Models that achieved the best validation accuracy were saved as the final ERM models.

\subsection{Training LBC Models}
To train our LBC models, we used a stochastic gradient descent optimizer with a  momentum of 0.9 and a weight decay of $10^{-4}$. The key hyperparameter settings and model selection criteria for training on the Waterbirds, CelebA, NICO, and ImageNet-9 datasets are shown in Table \ref{tab:lbc-training-details}. We used the ERM-trained models to initialize our LBC models. Standard data augmentations are used during training. The pseudo unbiased
validation accuracy exploits detected attributes and is defined in Section \ref{sec:training}. In each training epoch, we generated training batches by sampling with replacement in case we could not find enough samples under our within- and cross-class balancing techniques proposed in Section \ref{sec:training}.

\subsection{Time Costs for Extracting Attributes}
The VLM and the attribute extractor are only used once for offline data preparation on the training and validation splits of a dataset. The attribute extractor performs a single pass on the texts to find informative words with a linear time complexity. Thus, the overall time complexity wouldn't be a major concern, compared with costly human annotations. Specifically, the total processing time using ViT-GPT2 on a single NVIDIA RTX 8000 GPU for each of the datasets is in the table below.

\begin{table}[h]
\centering
\begin{tabular}{cc}
\hline
Datasets   & Time    \\ \hline
Waterbirds & 9.7min  \\
CelebA     & 4.6h    \\
NICO       & 16.2min \\
ImageNet-9 & 1.5h    \\ \hline
\end{tabular}
\caption{Time costs for extracting attributes from the four datasets.}
\end{table}

\begin{table*}[thbp]
\resizebox{\linewidth}{!}{%
\begin{tabular}{clcccccc}
\hline
Dataset                     & \multicolumn{1}{c}{Metric}                                                     & $\tanh(\text{Abs}(\log(\eta)))$ & $\tanh(\log(\eta))$ & $\text{Abs}(\log(\eta))$ & $\log(\eta)$ & $\text{Abs}(\delta)$ & $\delta$ \\ \hline
\multirow{3}{*}{Waterbirds} & \begin{tabular}[c]{@{}l@{}}Pseudo unbiased \\ validation accuracy\end{tabular} & \textbf{95.1}                                   & 94.3                       & 94.9                            & 95.0                & 94.7                      & 94.4          \\
                            & Average test accuracy                                                          & \textbf{93.2}                                   & 89.8                       & 91.3                            & 92.2                & 92.1                      & 91.0          \\
                            & Worst-group test accuracy                                                      & \textbf{87.3}                                   & 79.2                       & 82.9                            & 85.1                & 85.0                      & 81.7          \\ \hline
\multirow{3}{*}{CelebA}     & \begin{tabular}[c]{@{}l@{}}Pseudo unbiased \\ validation accuracy\end{tabular} & \textbf{94.6}                                   & 94.5                       & 94.3                            & 94.4                & 94.3                      & 94.3          \\
                            & Average test accuracy                                                          & 92.2                                   & 92.9                       & 93.0                            & \textbf{93.3  }              & 92.8                      & 93.3          \\
                            & Worst-group test accuracy                                                      & \textbf{81.2}                                   & 78.1                       & 79.1                            & 79.7                & 80.8                      & 78.8          \\ \hline
\end{tabular}%
}\caption{Comparison between different designs of spuriousness scores. We ran experiments using different scores for 5 times on the Waterbirds and CelebA datasets and calculated the average performance under different metrics.}\label{tab:spuriousness-score-compare}
\end{table*}

\section{Attributes with High Spuriousness Scores}
We give 10 spurious attributes with the highest spuriousness scores for each class of the CelebA, NICO, and ImageNet-9 datasets. As discussed in Section \ref{sec:spurious-detection} in the main paper, not all of these attributes are self-explanatory; some of them may represent features that cannot be described by themselves. In general, these spurious attributes  are not directly related to their corresponding classes.

\noindent\textbf{CelebA.}\\
\textit{Non-blond hair:} sun, umbrella, pretty, flag, lady, sky, blonde, ear, long, tooth\\
\textit{Blond hair:} apple, flag, right, animal, blow dryer, blow, dryer, bottle, hand, scarf.

\noindent\textbf{NICO.}\\
\textit{Dog}: ground, snow, white, green, lush, road, grassy, grass, sheep, side.\\
 \textit{Cat}: food, painting, snow, feeder, yellow, bird feeder, wood, colorful, parrot, seagull.\\
\textit{Bear:} grass, floor, animal, wire fence, bear, wire, person, hand, piece, branch.\\
\textit{Bird}: rock, bunch, grass, man, animal, bowl, large, room, chair, table.\\
\textit{Cow}: snow, beach, brown, top, woman, field, sandy, back, white, people.
\textit{Elephant}: leave, river, middle, tree, herd, man, stage, fence, body, baby elephant.\\
\textit{Horse}: group, banana, window, rock, picture, pile, people, plate, face, sign.\\
\textit{Monkey}: cat, beach, ground, hand, animal, snow, small, white, body, water.\\
\textit{Rat}: elephant, snow, man, body, herd, cow, cattle, water, field, white.\\
\textit{Sheep}: cement, bunch, plant, gray, wooden, banana, teddy, teddy bear, post, monkey.

\noindent\textbf{ImageNet-9.}  \\
\textit{Dog}: cell phone, phone, cell, right, desk, plant, cage, hand, picture, log.\\
\textit{Cat:} dirt road, statue, road, dirt, laptop, man, woman, bear, cat, large.\\
\textit{Frog}: woman, young, flower pot, boy, little, bunch, flower, girl, pot, body.\\
\textit{Turtle}: painting, boat, group, leave, dead, pile, animal, body, picture, beach.\\
\textit{Bird}: forest, mouth, middle, bird, white, duck, water, colorful, hummingbird, feeder\\
\textit{Primate}: trash can, trash, collage, can, parrot, flock, air, squirrel, dirt road, baby.\\
\textit{Fish}: fence, surfboard, shot, dog, right, person, fire hydrant, hydrant, fire, hand.\\
\textit{Crab}: view, beach scene, scene, flower, group, people, body, close, water, bunch.\\
\textit{Insect}: face, woman, front, knife, object, banana, hand, dog, person, animal.

\section{Different Designs of Spuriousness Score}
 We show the performance comparison of six variants of spuriousness score on the Waterbirds and CelebA datasets in Table \ref{tab:spuriousness-score-compare}, where $\delta=M(\mathcal{D}_{tr}^{( c,a)};f_{\theta})-M(\mathcal{D}_{tr}^{( c,\hat{a})};f_{\theta})$, $\eta=M(\mathcal{D}_{tr}^{( c,a)};f_{\theta})/M(\mathcal{D}_{tr}^{( c,\hat{a})};f_{\theta})$, and $M(\cdot;\cdot)$ is the accuracy measure used in Eq. \eqref{eq:spuriousness-score}. The models used for testing are selected based on the pseudo unbiased validation accuracy defined in Section \ref{sec:training}. We observe that taking the simple difference between the accuracies $M(\mathcal{D}_{tr}^{\langle c,a \rangle};f_{\theta})$ and $M(\mathcal{D}_{tr}^{\langle c,\hat{a} \rangle};f_{\theta}))$ is not as effective as taking the logarithm of their ratio. Therefore, adding non-linearity into our design of spuriousness score is beneficial. Moreover, $\tanh$ and $\text{Abs}$ together further improve the average and worst-group test accuracies of our proposed method on the Waterbirds dataset. On the CelebA dataset, the default score, i.e., $\tanh(\text{Abs}(\log(\eta)))$, achieves the best pseudo unbiased validation accuracy, which favors a model that achieves the best worst-group test accuracy. Overall, our spuriousness score works well with the pseudo unbiased validation accuracy in selecting a model that is most robust to spurious correlations in terms of worst-group test accuracy and has competitive average performance.

\section{Analysis Based on Spuriousness Score}
\begin{figure}[htbp]
    \centering
    \includegraphics[width=\linewidth]{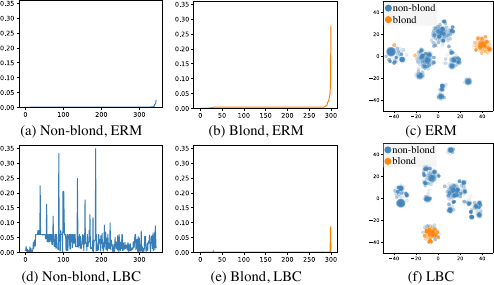}
    \caption{(a) and (b): Spuriousness scores for the attributes detected from non-blond and blond based on an ERM model. (d) and (e): Spuriousness scores based on our LBC model. (c) and (f): Spurious embeddings of the images in the CelebA dataset based on the ERM and LBC model, respectively.}
    \label{fig:spurousness-score-celeba}
\end{figure}

We additionally show the spuriousness scores of the attributes detected within the non-blond and blond classes in the CelebA dataset before (denoted as ERM) and after applying our proposed LBC. The high maximum score in Fig. \ref{fig:spurousness-score-celeba}(b) shows that for the ERM model, predicting the blond class heavily relies on spurious correlations, while predicting the non-blond class is relatively robust to spurious correlations as the maximum score in Fig. \ref{fig:spurousness-score-celeba}(a) is small. This also aligns with our empirical observation that the ERM model struggles in predicting the blond class. Fig. \ref{fig:spurousness-score-celeba}(c) shows that some of the prediction behaviors (orange points) for predicting images from the blond class are similar to those leading to the non-blond class, offering insights into why the ERM model performs poorly on predicting the blond class.

After our LBC retraining, as shown in Fig. \ref{fig:spurousness-score-celeba}(e), the reliance on spurious correlations is significantly reduced. However, as a side effect, the reliance on spurious correlations increases for predicting the non-blond class, as observed in Fig. \ref{fig:spurousness-score-celeba}(d). As a result, for images in the non-blond class, we observe dense clusters in Fig. \ref{fig:spurousness-score-celeba}(f) with each cluster representing similar prediction behaviors which use certain spurious correlations for predictions. Interestingly, we observe that images in the blond class are more concentrated in the spuriousness embedding space after our LBC retraining, indicating more consistent prediction behaviors on the class. This improved consistency comes at the cost of increased inconsistency in the predictions of the non-blond class images, as we observe that several non-blond class images (blue points) are close to the orange cluster. Given that the non-blond class is the majority class, while the blond class is the minority class, the increased consistency in predicting blond class images improves the performance on the blond class images reflected by the increased worst-group accuracy. At the same time, the average accuracy dominated by the non-blond class images decreases due to the increased inconsistency in the predictions of the non-blond class images. This average and worst-group accuracy tradeoff is commonly observed in Table \ref{tab:celeba-waterbirds} in the main paper across different methods, and our spuriousness score can effectively reveal this tradeoff.

\section{Analysis on Using ERM-Trained Models}
Our method starts training using the initialization of an ERM-trained model. To investigate how different initializations affect the performance of our method, we tested three kinds of models used by our method: (1) a randomly initialized model, (2) an ERM model trained from scratch, and (3) an ERM model trained with ImageNet pre-trained weights. Table \ref{tab:model-initialization-comparison} shows that LBC with a randomly initialized model does not perform well on the three evaluation metrics, because the randomly initialized model gives noisy information on the spuriousness of the detected attributes. LBC with an ERM model trained from scratch performs better than the first one thanks to the good initialization provided by the ERM-trained model. The ImageNet pre-trained weights contain knowledge about recognizing multiple objects and patterns. Therefore, when the ERM model is trained with ImageNet pre-trained weights, LBC performs the best on the three metrics.

\begin{table}[h]
\resizebox{\linewidth}{!}{%
\begin{tabular}{ccccc}
\hline
\multirow{2}{*}{Method} & \multirow{2}{*}{ERM Model}                                                           & \multicolumn{2}{c}{ImageNet-9}                  & ImageNet-A        \\ \cline{3-5} 
                        &                                                                                      & Validation $(\uparrow)$ & Unbiased $(\uparrow)$ & Test $(\uparrow)$ \\ \hline
LBC                     & Random initialization                                                                & 46.38                 & 43.92                & 15.73             \\
LBC                     & Trained from scratch                                                                 & 93.71                   & 92.14                 & 39.65             \\
LBC                     & \begin{tabular}[c]{@{}c@{}}Trained with ImageNet \\ pre-trained weights\end{tabular} & 96.97                   & 96.03                 & 40.63             \\ \hline
\end{tabular}%
}
\caption{Performance comparison (\%) between different choices of model initializations used in our method LBC on the ImageNet-9 and ImageNet-A datasets.}\label{tab:model-initialization-comparison}
\end{table}

\section{Does the Performance Gain Come from the Attributes?}
Since we used a VLM to detect attributes from training data, it is naturally to ask whether the performance gain comes from the detected attributes. We showed that the performance gain mainly comes from our proposed learning algorithm. Specifically, we added an \textit{additional} layer after the backbone to predict attributes for each image, and we trained the whole model on the Waterbirds and CelebA datasets, respectively. In other words, we added an additional attribute prediction loss term in Eq. \eqref{eq:erm-objective} for each image. Essentially, the attributes act as a regularization for the classifier. If the attributes contain information effective in improving a classifier's robustness to spurious correlations, we would observe improved performance after training.

The worst-group accuracies on the Waterbirds and CelebA datasets are 71.7\% and 47.2\%, respectively. 
Although this approach is slightly better than ERM, but it falls far behind our proposed LBC algorithm. Therefore, the detected attributes from the VLM alone do not contain information effective for improving a classifier's robustness to spurious correlations. In contrast, LBC directly identifies highly dependent spurious attributes for a classifier and mitigates the classifier's reliance on them, effectively improving the classifier's robustness to spurious correlations.  
\end{document}